\documentclass[lettersize,journal,twoside]{IEEEtran}
\IEEEoverridecommandlockouts    % This command is only needed if
\usepackage{booktabs}
\usepackage{graphicx}
\usepackage[colorlinks,linkcolor=red,anchorcolor=blue,citecolor=green]{hyperref}
\usepackage{cite}
\usepackage{array}
\usepackage{multirow}
\usepackage{tabularx} 
\usepackage{arydshln}
\usepackage{makecell}
\usepackage[font=small]{caption}
\usepackage{graphicx} % for pdf, bitmapped graphics files
\usepackage{amsmath} % assumes amsmath package installed
\usepackage{amssymb}  % assumes amsmath package installed

\usepackage[capitalize]{cleveref}
\crefname{section}{Sec.}{Secs.}
\Crefname{section}{Section}{Sections}
\Crefname{table}{Table}{Tables}
\crefname{table}{Tab.}{Tabs.}
\usepackage[table,xcdraw]{xcolor}
\usepackage{CJKutf8}

\newcommand{\name}[0]{TVG-SLAM}

\title{\LARGE \bf
\name{}: Robust Gaussian Splatting SLAM with Tri-view Geometric Constraints
}

\author{
Zhen Tan, Xieyuanli Chen, Lei Feng, Yangbing Ge, Shuaifeng Zhi, Jiaxiong Liu, Dewen Hu*
\thanks{All authors are with the National University of Defense Technology, China.}%
\thanks{* indicates corresponding author: D. Hu (dwhu@nudt.edu.cn)}
}

\begin{document}

\maketitle
\thispagestyle{empty}
\pagestyle{empty}

%%%%%%%%%%%%%%%%%%%%%%%%%%%%%%%%%%%%%%%%%%%%%%%%%%%%%%%%%%%%%%%%%%%%%%%%%%%%%%%%
\begin{abstract}
Recent advances in 3D Gaussian Splatting (3DGS) have enabled RGB-only SLAM systems to achieve high-fidelity scene representation. 
However, the heavy reliance of existing systems on photometric rendering loss for camera tracking undermines their robustness, especially in unbounded outdoor environments with severe viewpoint and illumination changes.
To address these challenges, we propose TVG-SLAM, a robust RGB-only 3DGS SLAM system that leverages a novel tri-view geometry paradigm to ensure consistent tracking and high-quality mapping.
We introduce a dense tri-view matching module that aggregates reliable pairwise correspondences into consistent tri-view matches, forming robust geometric constraints across frames.
For tracking, we propose Hybrid Geometric Constraints, which leverage tri-view matches to construct complementary geometric cues alongside photometric loss, ensuring accurate and stable pose estimation even under drastic viewpoint shifts and lighting variations.
For mapping, we propose a new probabilistic initialization strategy that encodes geometric uncertainty from tri-view correspondences into newly initialized Gaussians.
Additionally, we design a Dynamic Attenuation of Rendering Trust mechanism to mitigate tracking drift caused by mapping latency. 
Experiments on multiple public outdoor datasets show that our TVG-SLAM outperforms prior RGB-only 3DGS-based SLAM systems.
Notably, in the most challenging dataset, our method improves tracking robustness, reducing the average Absolute Trajectory Error (ATE) by 69.0\% while achieving state-of-the-art rendering quality.
The implementation of our method will be released as open-source.
\end{abstract}

% Hybrid Geometric Constraints
% probabilistic initialization strategy
% Dynamic Attenuation of Rendering Trust (DART)

%%%%%%%%%%%%%%%%%%%%%%%%%%%%%%%%%%%%%%%%%%%%%%%%%%%%%%%%%%%%%%%%%%%%%%%%%%%%%%%%
\section{INTRODUCTION}
Simultaneous Localization and Mapping (SLAM) is a cornerstone technology for robotics, autonomous driving, and augmented reality \cite{cadena2016past}. 
The evolution of SLAM has seen a paradigm shift from traditional methods \cite{newcombe2011dtam, salas2013slam++, mur2017orb2, campos2021orb3, whelan2015elasticfusion, newcombe2011kinectfusion, bloesch2018codeslam, schops2019bad, ge2024pipo} to neural rendering-based approaches \cite{sucar2021imap, zhu2022nice, wang2021neus, chung2023orbeez, wang2023co, zhang2023go, li2023end, guo2024neuv, zhu2024nicer}. 
Recently, 3D Gaussian Splatting (3DGS) \cite{kerbl2023gaussians} has significantly advanced the field. 
By representing scenes with explicit Gaussian primitives, 3DGS enables photorealistic real-time rendering alongside high-fidelity mapping, inspiring novel 3DGS-based SLAM frameworks \cite{yan2023gs, yugay2023gaussianslam, keetha2023splatam, matsuki2023monogs, huang2023photo, wang2024opengs}.

Despite promising results, current 3DGS-based SLAM systems face a fundamental challenge: their camera tracking depends heavily on photometric consistency between the current image and rendered views~\cite{yugay2023gaussianslam, matsuki2023monogs, engel2018dso}. 
This assumption, inherited from early NeRF-based \cite{mildenhall2021nerf} SLAM methods like iMAP \cite{sucar2021imap} and NICE-SLAM \cite{zhu2022nice}, is prone to failure under real-world conditions involving rapid motion, lighting changes, or texture-less areas. 
This fragility is especially pronounced for RGB-only systems in unbounded outdoor environments, where dynamic illumination (e.g., shadows, clouds, varying sun angles) and large viewpoint shifts critically hinder robustness. Additionally, existing mapping methods rely on heuristic Gaussian initialization, often leading to geometric inaccuracies and suboptimal scene representations.

\begin{figure}[t]
  \centering
  \includegraphics[width=\linewidth]{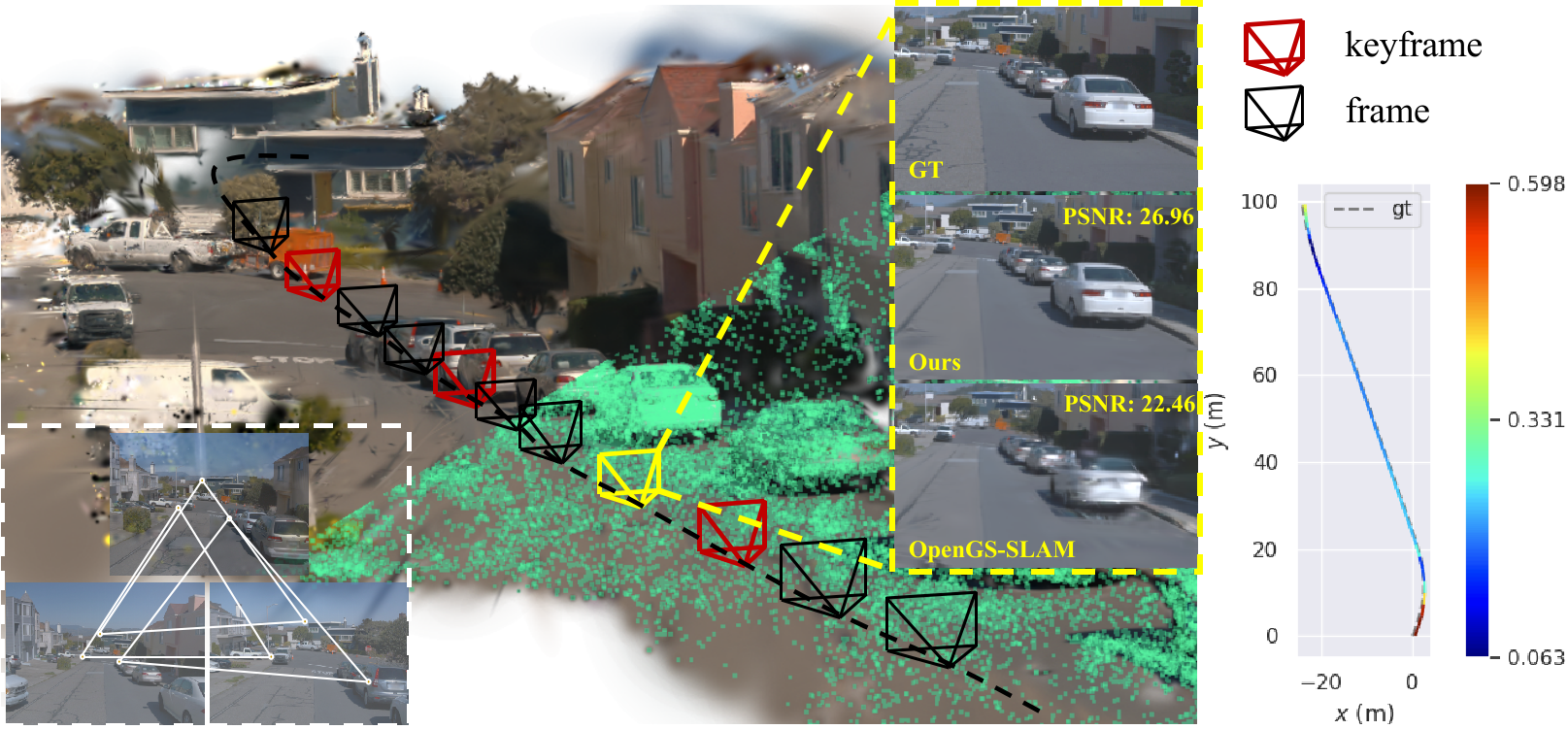}
  \caption{Our system integrates tri-view geometric constraints to achieve both high-fidelity representation and a highly accurate camera trajectory that closely aligns with the ground truth.}
  \label{fig:teaser}
  % \vspace{-0.2cm}
\end{figure}

To address the challenges faced by RGB-only SLAM systems in outdoor environments, we propose TVG-SLAM (as shown in~\cref{fig:teaser}), a novel system centered around a dense tri-view geometry paradigm for reliable tracking and high-fidelity mapping. At the core of our approach is a dense tri-view matching module, which constructs consistent correspondences across three consecutive frames by aggregating pairwise matches. This tri-view association significantly enhances the stability and reliability of multi-view geometric constraints.

Building on dense tri-view matching, our tracking module introduces Hybrid Geometric Constraints, which combine photometric loss with trifocal-based 2D reprojection and 3D alignment losses to ensure robust pose estimation under challenging conditions. For mapping, we propose TUGI, a probabilistic initialization strategy that uses multi-view geometric uncertainty to guide Gaussian initialization. To further enhance stability, we introduce DART, which dynamically downweighs photometric cues when mapping lags, mitigating drift. Together, these components form a cohesive SLAM system that tightly integrates photometric and geometric constraints. TVG-SLAM outperforms prior RGB-only 3DGS-based methods in both tracking accuracy and rendering fidelity across challenging outdoor benchmarks.

The main contributions are listed as follows:
\begin{itemize}
    \item We first introduce the tri-view geometry paradigm into the GS-based SLAM framework, enabling robust data association and geometric reasoning across frames.
    \item We design Hybrid Geometric Constraints that jointly leverage photometric consistency, trifocal 2D reprojection, and 3D alignment losses, improving pose robustness under large viewpoint and illumination changes.
    \item We propose TUGI, a probabilistic Gaussian initialization strategy that encodes multi-view geometric uncertainty into the Gaussian parameters, improving map quality and rendering fidelity.
    \item We develop DART, a dynamic weighting mechanism that attenuates photometric loss supervision when the map becomes stale, mitigating tracking drift in asynchronous SLAM scenarios.
    % \item We propose TUGI, a probabilistic mapping strategy that uses tri-view geometric uncertainty to guide Gaussian initialization and develop DART that adaptively reduces reliance on photometric cues during mapping delays, mitigating tracking drift.
    % \item We demonstrate state-of-the-art performance on challenging outdoor datasets, achieving superior accuracy in both pose estimation and rendering quality compared to existing RGB-only GS SLAM systems.
\end{itemize}

% \begin{figure*}[t!]
%     \centering
%     \includegraphics[width=0.95\linewidth]{figs/overview.png}
%     \caption{The \name{} pipeline. Our system processes incremental RGB images by first building robust tri-view matches. In tracking, we utilizes Hybrid Geometric Constraints—jointly optimizing photometric, trifocal, and 3D alignment losses—to estimate the camera pose robustly. In mapping, keyframes are integrated into the map using our Tri-view Uncertainty-guided Gaussian Initialization (TUGI) strategy, which leverages multi-view geometric consistency to initialize new Gaussians for a high-fidelity scene representation.}
%     \vspace{-0.3cm}
%     \label{fig:overview}
% \end{figure*}
\section{RELATED WORK}
\label{sec:related}
\subsection{Pose Optimization in NeRF and 3DGS}
Neural rendering methods such as NeRF \cite{mildenhall2021nerf} and 3DGS \cite{kerbl2023gaussians} achieve high-quality scene reconstruction but typically rely on known or externally estimated camera poses. To remove this dependency, recent works have proposed jointly optimizing poses and scene parameters. Early NeRF-based works like NeRF-- \cite{wang2021nerfmm} and BARF \cite{lin2021barf} pioneered this joint optimization for static scenes. Subsequent methods \cite{jeong2021scnerf, chen2023l2g, bian2022nope, tan2024td, yuan2024ranerf, ran2024ctnerf} further improved robustness for complex camera trajectories, but these approaches are often limited to offline settings and may require sparse priors or global optimization.
Other methods, including CF-NeRF \cite{yan2024cfnerf} and CF-3DGS \cite{fu2024cfgs}, aim to eliminate explicit pose estimation entirely by leveraging dense correspondences or learned feature alignment. While promising, these systems are still designed for static, offline reconstruction and lack support for online tracking, incremental mapping, or spatiotemporal consistency.

\subsection{Neural Rendering SLAM}
To bridge this gap, prior works have integrated neural scene representations into SLAM pipelines, aiming to enable online tracking and mapping.
Early neural SLAM systems~\cite{sucar2021imap, zhu2022nice, wang2023co, zhang2023go, yang2022vox, liso2024loopy, zhu2024nicer} incorporated neural radiance fields (NeRF)~\cite{mildenhall2021nerf} into their mapping and tracking pipelines, enabling photorealistic scene reconstruction and joint pose optimization. However, the implicit nature of NeRF leads to slow rendering and expensive optimization, making it less suitable for online SLAM applications.

Recent works~\cite{keetha2023splatam, yugay2023gaussianslam, yan2023gs} adopt 3DGS~\cite{kerbl2023gaussians}, an explicit scene representation that allows differentiable rendering and fast gradient-based updates. 3DGS-based SLAM methods typically rely on RGB-D inputs to ensure stable tracking and accurate mapping, leveraging depth supervision to reduce ambiguity.
To further improve efficiency and reduce hardware requirements, some approaches~\cite{matsuki2023monogs, huang2023photo, sandstrom2025splat, zhu2024mgs, guo2024motiongs, wang2024opengs} have extended 3DGS-based SLAM to RGB-only settings. While these methods demonstrate promising results, they heavily depend on photometric rendering loss for tracking. This makes them sensitive to lighting changes, large viewpoint shifts, and texture-less regions—issues commonly encountered in unbounded outdoor environments.

To address these challenges, we propose TVG-SLAM, an RGB-only SLAM system built upon a tri-view geometry paradigm that introduces dense geometric constraints and uncertainty-guided Gaussian initialization, improving robustness and accuracy in outdoor scenes.

\begin{figure*}[t!]
    \centering
    \includegraphics[width=0.95\linewidth]{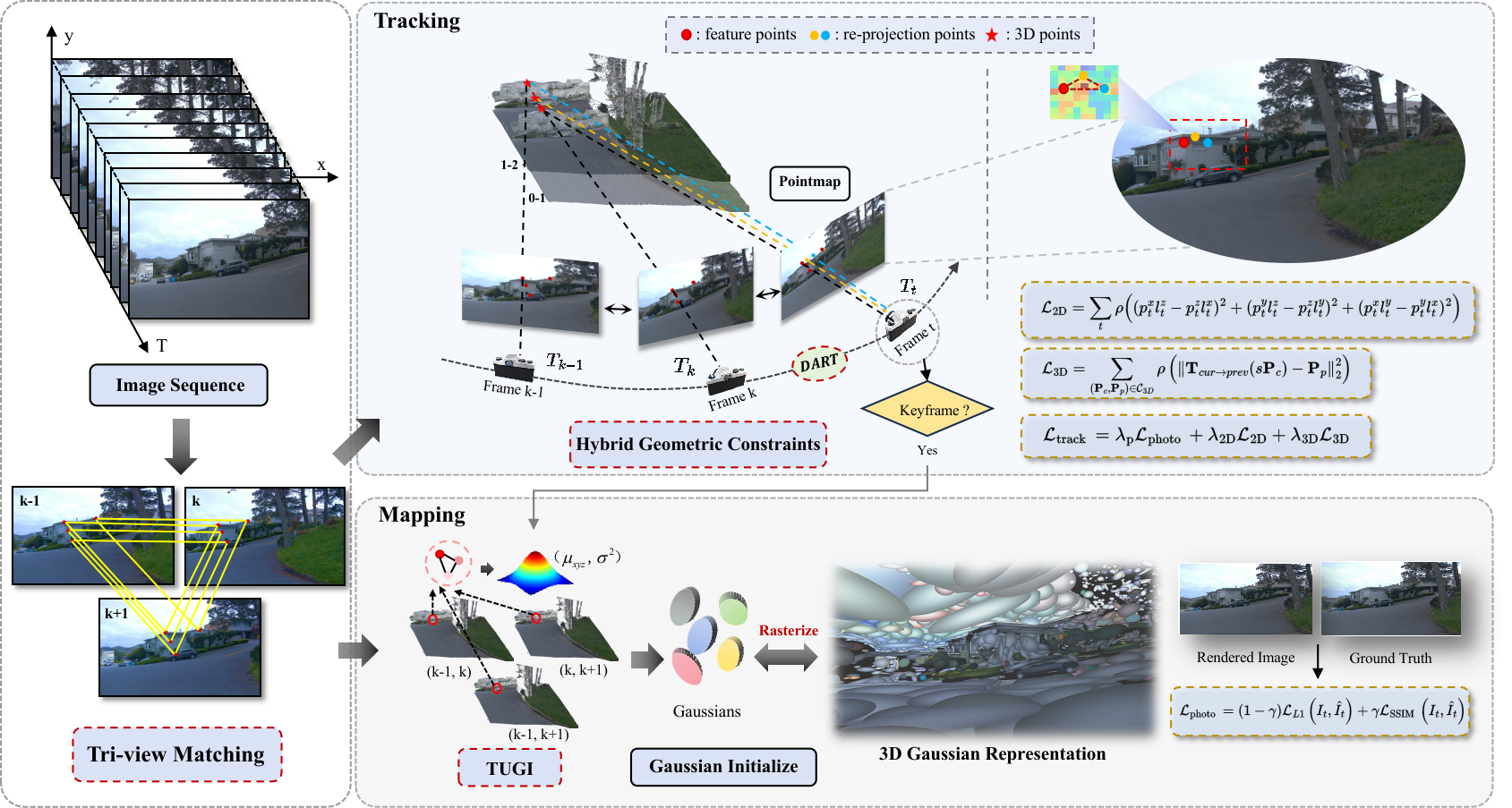}
    \caption{The \name{} pipeline. Our system processes incremental RGB images by first building robust tri-view matches. In tracking, we utilizes Hybrid Geometric Constraints—jointly optimizing photometric, trifocal, and 3D alignment losses—to estimate the camera pose robustly. In mapping, keyframes are integrated into the map using our Tri-view Uncertainty-guided Gaussian Initialization (TUGI) strategy, which leverages multi-view geometric consistency to initialize new Gaussians for a high-fidelity scene representation.}
    \vspace{-0.2cm}
    \label{fig:overview}
\end{figure*}

\section{METHOD}
\label{sec:method}
% This section introduces \name{}, our novel RGB-only SLAM system, addressing outdoor robustness challenges by integrating a tri-view geometry paradigm into both tracking and mapping. This significantly enhances camera pose estimation and mapping fidelity. The overall framework is illustrated in~\cref{fig:overview} and detailed below.

\subsection{System Overview}
As illustrated in \cref{fig:overview}, \name{} is an RGB-only SLAM system designed for challenging outdoor environments. Its pipeline consists of three tightly integrated components that couple geometric reasoning with photometric rendering.
First, a dense matching module builds reliable \textbf{tri-view correspondences} across frames. 
These are fed into a \textbf{hybrid geometric tracking} module that jointly optimizes photometric, 2D reprojection, and 3D alignment losses. This module features our DART mechanism to mitigate drift from mapping delays by adaptively reducing reliance on photometric cues. 
Finally, an \textbf{uncertainty-guided mapping} module incrementally reconstructs the 3D Gaussian map, initializing new Gaussians with uncertainty-aware priors derived from multi-view consistency.

\subsection{Preliminaries: 3D Gaussian Representation}
Our system uses 3DGS \cite{kerbl2023gaussians} for scene representation. Unlike implicit representations like NeRF, 3DGS models a scene using a set of explicit, interpretable primitives: anisotropic 3D Gaussians. 
Each Gaussian is defined by a mean (position) $\boldsymbol{\mu} \in \mathbb{R}^3$, a covariance matrix $\boldsymbol{\Sigma} \in \mathbb{R}^{3\times3}$, a color $\mathbf{c} \in \mathbb{R}^3$ (stored as spherical harmonic coefficients), and an opacity $\alpha \in \mathbb{R}$. The covariance matrix $\boldsymbol{\Sigma}$, which describes the shape and orientation, is decomposed into a scaling matrix $\mathbf{S}$ and a rotation matrix $\mathbf{R}$ ($\boldsymbol{\Sigma} = \mathbf{R}^T\mathbf{S}^T\mathbf{S}\mathbf{R}$) for efficient optimization.

To render a 2D image, 3DGS employs an efficient, differentiable pipeline. For a given camera pose $\mathbf{T}$, the 3D mean $\boldsymbol{\mu}$ is projected into the camera's coordinate system. The 3D covariance $\boldsymbol{\Sigma}$ is then projected into a 2D covariance $\boldsymbol{\Sigma}'$ via an affine approximation. Finally, the color $\mathbf{C}(\mathbf{p})$ at any pixel $\mathbf{p}$ is computed by alpha-blending all $N$ sorted Gaussians along the camera ray:
\begin{equation}
\mathbf{C}(\mathbf{p}) = \sum_{i=1}^{N} \mathbf{c}_i \alpha_i \prod_{j=1}^{i-1}(1 - \alpha_j),
\label{eq:render}
\end{equation}
where $\alpha_i$ is the $i$-th Gaussian's opacity. The entire process is differentiable, including Gaussian parameters and the camera pose $\mathbf{T}$, enabling joint optimization via gradient descent.

\subsection{Dense Tri-view Matching}
Reliable data association is essential for introducing geometric constraints into SLAM systems. The core of our method is a dense tri-view matching strategy that establishes high-quality multi-view correspondences across the current frame and two nearby keyframes, forming the basis for the geometric objectives in both our tracking and mapping modules.

Specifically, we use the deep dense matcher DUST3R~\cite{wang2024dust3r} to compute pairwise correspondences between the current incoming frame $I_t$ and the two most recent keyframes $I_k$ and $I_{k-1}$. Here, $I_k$ and $I_{k-1}$ denote the latest and previous keyframes, while $I_t$ is the frame currently being tracked (not yet determined as a keyframe).

To construct temporally consistent triplet matches, we adopt a bridging strategy centered at $I_k$:
\begin{enumerate}
    \item Compute two dense match sets: $M_{k,t}$ between $I_k$ and $I_t$, and $M_{k-1,k}$ between $I_{k-1}$ and $I_k$, each consisting of pixel-level correspondences and their associated 3D pointmaps.
    \item For each match $(\mathbf{p}_k, \mathbf{p}_t)$ in $M_{k,t}$, search for a corresponding match $(\mathbf{p}_{k-1}, \mathbf{p}_k)$ in $M_{k-1,k}$ that shares the pixel $\mathbf{p}_k$ in the intermediate keyframe.
    \item A successful pair yields a tri-view correspondence $(\mathbf{p}_{k-1}, \mathbf{p}_k, \mathbf{p}_t)$. All such triplets form the tri-view match set $\mathcal{M}_{k-1,k,t}$.
\end{enumerate}

This process enforces temporal consistency and improves match reliability by filtering out transient or inconsistent correspondences that may appear in individual pairs. Moreover, each triplet offers at least two independent pointmaps, enabling geometric redundancy that enhances the robustness of subsequent tracking and mapping stages.

\subsection{Hybrid Geometric Tracking}
Given a set of reliable tri-view correspondences $\mathcal{M}_{k-1,k,t}$, we formulate a hybrid objective to estimate the 6-DoF pose $\mathbf{T}_{t} \in SE(3)$ of the current tracking frame $I_{t}$.
Our core tri-view paradigm enables the formulation of multiple geometric constraints. Specifically, we design a loss function that integrates photometric consistency with two complementary geometric terms derived from tri-view matches: a 2D constraint based on the classical trifocal tensor and a direct 3D alignment constraint. The total loss is:
\begin{equation}
\mathcal{L}_{\text{track}} = \lambda_{\text{p}} \mathcal{L}_{\text{photo}} + \lambda_{2\text{D}} \mathcal{L}_{2\text{D}} + \lambda_{3\text{D}} \mathcal{L}_{3\text{D}},
\label{eq:total_loss}
\end{equation}
where $\lambda_{\text{p}}, \lambda_{2\text{D}}, \lambda_{3\text{D}}$ are the weights for the photometric, 2D trifocal, and 3D alignment losses, respectively.

\textbf{Photometric Loss:} 
$\mathcal{L}_{\text{photo}}$ measures the similarity between the observed image $I_{t}$ and the rendered view $\hat{I}_{t}$ from the 3D Gaussian map using the estimated pose $\mathbf{T}_{t}$. We use a combination of L1 and SSIM losses:
\begin{equation}
\mathcal{L}_{\text{photo}} = (1-\gamma) \mathcal{L}_{L1}(I_{t}, \hat{I}_{t}) + \gamma  \mathcal{L}_{\text{SSIM}}(I_{t}, \hat{I}_{t}),
\end{equation}
where $\gamma$ is a fixed hyperparameter.

%%%%%%%%%%%%%%%%%%%%%%%%%%%%%%%%%%%%%%%%%%%%%%%%%%%%%%%%%%%%%%
\textbf{Trifocal Constraint Loss:} 
$\mathcal{L}_{\text{{2D}}}$ enforces a pure multi-view geometric constraint based on the trifocal tensor~\cite{Hartley2004mvg}, without requiring explicit 3D reconstruction. For each tri-view correspondence $(\mathbf{p}_{k-1}, \mathbf{p}_{k}, \mathbf{p}_{t})$, where $\mathbf{p}_{k-1}$ and $\mathbf{p}_{k}$ are matched points in two keyframes and $\mathbf{p}_t$ is the corresponding point in the current frame, we first compute the trifocal tensor $\mathcal{T}$ from the relative camera poses. The trifocal tensor consists of three $3\times 3$ matrices $\mathcal{T} = \{ \mathcal{T}_1, \mathcal{T}_2, \mathcal{T}_3 \}$, each corresponding to one component of the point $\mathbf{p}_{k-1} = (p_{k-1}^1, p_{k-1}^2, p_{k-1}^3)^\top$.

Using the epipolar line transfer formula, the corresponding epipolar line $\mathbf{l}_t \in \mathbb{R}^3$ in the third view is computed as:
\begin{equation}
\mathbf{l}_t = \sum_{i=1}^{3} p_{k-1}^i \mathcal{T}_i \mathbf{p}_{k},
\end{equation}
We then define the loss as the sum of squared algebraic residuals, which approximate the geometric distance between the point $\mathbf{p}_t = (p_t^x, p_t^y, p_t^z)$ and the epipolar line $\mathbf{l}_t = (l_t^x, l_t^y, l_t^z)$:
\begin{equation}
\mathcal{L}_{\text{2D}} = \sum_t \rho \Big(
    (p_t^x l_t^z - p_t^z l_t^x)^2 +
    (p_t^y l_t^z - p_t^z l_t^y)^2 +
    (p_t^x l_t^y - p_t^y l_t^x)^2
\Big),
\end{equation}
where $\rho(\cdot)$ is a Huber loss function to reduce the influence of outliers. In essence, this loss minimizes the geometric distance between an observed point in the current frame and its corresponding epipolar line, which is determined by the other two views and their relative poses. 

Trifocal constraints offer stronger geometric observability than pairwise epipolar geometry. As analyzed in~\cite{indelman2012iLBA}, they remain effective under degenerate motions like collinear trajectories, making them especially suitable for outdoor SLAM with low parallax and straight-line motion.

This strategy offers a strong map-independent constraint for robust tracking, especially in challenging regions where photometric consistency is unreliable. To ensure robustness, we only use matches that satisfy favorable geometric conditions (e.g., sufficient parallax) to avoid degenerate configurations.

\textbf{3D Alignment Loss:} 
$\mathcal{L}_{\text{3D}}$ provides a non-projective geometric constraint directly in 3D space. 
Instead of relying on traditional triangulation, we leverage pointmaps generated by~\cite{wang2024dust3r}.
For a tri-view match $(\mathbf{p}_{k-1}, \mathbf{p}_k, \mathbf{p}_{t})$, we can obtain two corresponding 3D points from the two pairwise matches: $\mathbf{P}_{c}$ from the $(I_k, I_{t})$ pointmap (in the current frame's coordinates) and $\mathbf{P}_{p}$ from the $(I_{k-1}, I_k)$ pointmap (in the previous frame's coordinates). 
Ideally, these points should align in 3D space after the correct transformation. 
We first estimate a relative scale factor $s$ and then compute the transformation $\mathbf{T}_{cur \rightarrow prev}$ from the current to the previous frame. 
The 3D alignment loss is defined as:
\begin{equation}
\mathcal{L}_{\text{3D}} = \sum_{(\mathbf{P}_{c}, \mathbf{P}_{p}) \in \mathcal{C}_{3D}} \rho \left( \left\| \mathbf{T}_{cur \rightarrow prev} (s \mathbf{P}_{c}) - \mathbf{P}_{p} \right\|_2^2 \right),
\label{eq:3d_loss}
\end{equation}
where $\mathcal{C}_{3D}$ is the set of all valid 3D-3D correspondences.
The scale factor $s$ is estimated using Procrustes alignment~\cite{umeyama1991least} between the two point sets.

\textbf{Dynamic Attenuation of Rendering Trust (DART):}
The relative weights in Eq.~\ref{eq:total_loss} determine the influence of different loss terms on pose optimization. 
However, fixed weights fail to adapt to the dynamic nature of SLAM, especially under asynchronous tracking and mapping. 
When the mapping thread lags—e.g., during aggressive motion—the rendered view used for $\mathcal{L}_{\text{photo}}$ may be based on an outdated map, degrading its reliability. 
Over-reliance on this inaccurate photometric supervision can lead to erroneous pose updates.

To address this, we propose DART, a dynamic re-weighting strategy that modulates the influence of $\mathcal{L}_{\text{photo}}$ based on the freshness of the underlying 3D map. 
Specifically, we use the number of frames processed since the last keyframe, denoted as $\Delta N_f$, as a proxy for map staleness. A larger $\Delta N_f$ implies the rendered view is less reliable.

We model the photometric loss weight $\lambda_{\text{p}}$ as a decreasing sigmoid-like function of $\Delta N_f$, ensuring a smooth and continuous attenuation:
\begin{equation}
\lambda_{\text{p}} = w_{\min} + (w_{\max} - w_{\min}) \sigma(k(\Delta N_f - N_m)),
\label{eq:dart}
\end{equation}
where $\sigma(x) = 1/(1+e^x)$ is the sigmoid function, $w_{\min}$ and $w_{\max}$ denote the minimum and maximum photometric weights, $N_m$ is the midpoint, and $k$ controls the sharpness of the transition.

With DART, when the map is recently updated (small $\Delta N_f$), $\lambda_{\text{p}}$ remains high, leveraging accurate photometric supervision. As the map becomes stale, $\lambda_{\text{p}}$ smoothly decays, allowing the system to rely more on our robust, map-independent geometric constraints ($\mathcal{L}_{2D}$ and $\mathcal{L}_{3D}$). This adaptive, trust-aware reweighting mechanism enhances tracking robustness in rapidly changing environments.

\subsection{Uncertainty-Guided Mapping}
During mapping, our system incrementally builds a globally consistent and geometrically accurate 3D Gaussian map using newly selected keyframes. Existing methods often use heuristics (e.g., based on photometric gradients or depth residuals) to decide how to initialize new Gaussians. This approach underutilizes the rich information available from tri-view geometry and can lead to inaccurate or redundant Gaussians in poorly constrained regions. We therefore propose TUGI, a principled Gaussian initialization strategy guided by tri-view geometric variance.

\subsubsection{Uncertainty Estimation from Tri-view Consistency}
To assess the geometric reliability of new candidate points, we estimate their 3D uncertainty directly from tri-view matches. Given a triplet of frames $(I_{k-1}, I_k, I_{k+1})$, dense pairwise correspondences provide two or more independent 3D estimates of the same scene point via pointmaps.
For each tri-view correspondence $(\mathbf{p}_{k-1}, \mathbf{p}_k, \mathbf{p}_{k+1})$, we retrieve its associated 3D positions from the matched pointmaps: $\mathbf{P}_{k \leftarrow k-1}$ from the $(I_{k-1}, I_k)$ pair, $\mathbf{P}_{k \leftarrow k+1}$ from $(I_k, I_{k+1})$, and $\mathbf{P}_{k \leftarrow k+1 \rightarrow k-1}$ from $(I_{k+1}, I_{k})$ reprojected into the same coordinate frame. These 3D points are then transformed into a common reference frame (typically $I_k$ or $I_{k-1}$) using known relative poses.

We define the uncertainty score $\sigma_g^2$ as the isotropic variance of these estimates:
\begin{equation}
\sigma_g^2=\frac{1}{N} \sum_{i=1}^N\left\|\mathbf{P}_i-\overline{\mathbf{P}}\right\|_2^2,
\end{equation}
where $\bar{\mathbf{P}}$ is the centroid of the valid 3D positions and $N \geq 2$ is the number of valid estimates. This score captures the multi-view consistency of the point: smaller $\sigma_g^2$ indicates high agreement and low geometric uncertainty.

Unlike prior heuristics based on photometric gradients or depth confidence, this formulation leverages direct multi-view geometric evidence. The estimated $\sigma_g^2$ is used to guide the initialization of Gaussian primitives, enabling uncertainty-aware map building.

\begin{figure*}[t]
  \centering
  \includegraphics[width=0.98\linewidth]{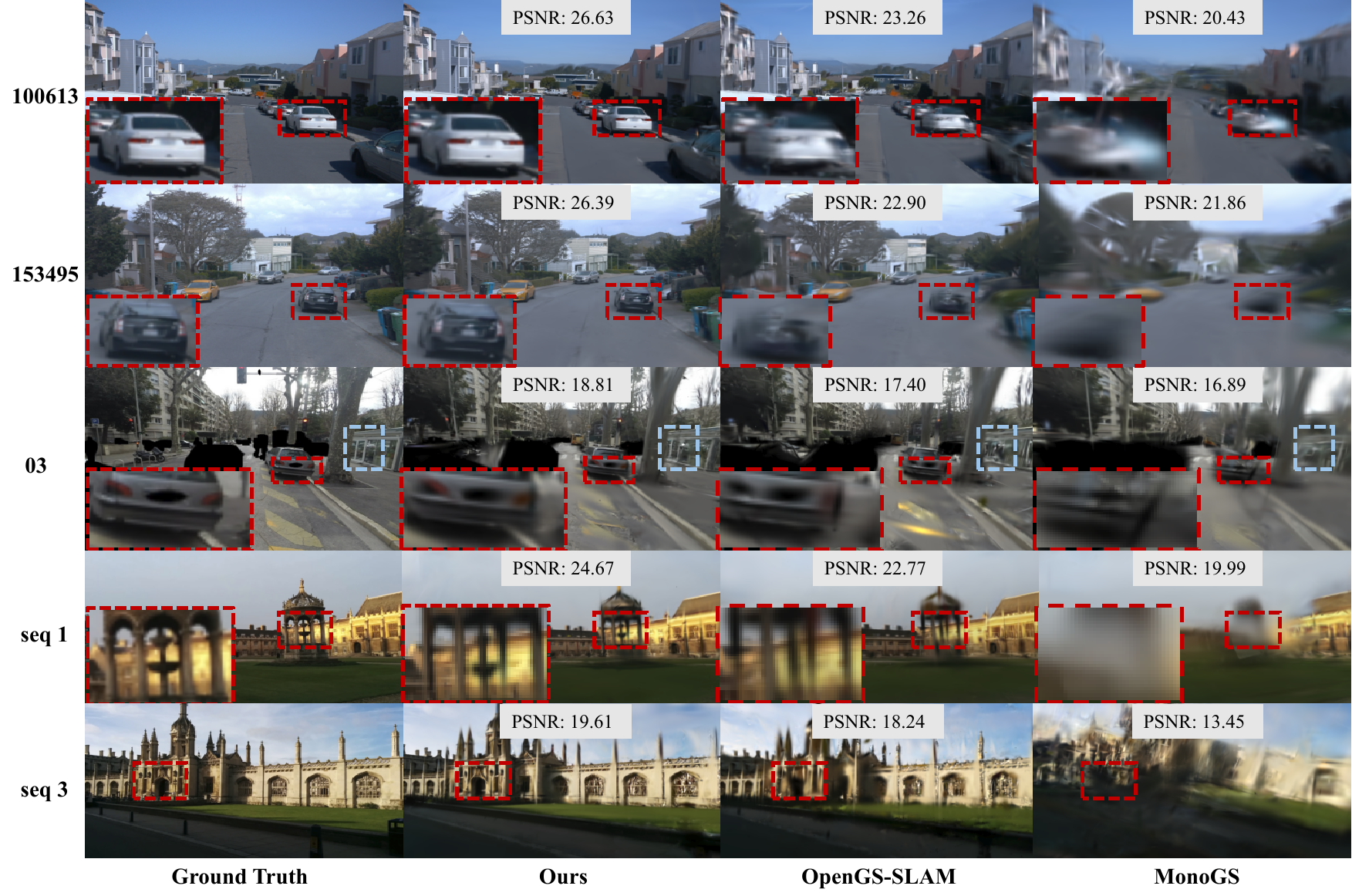}
  % \captionsetup{aboveskip=0pt, belowskip=0pt}
  \vspace{-0.2cm}
  \caption{Rendering quality comparison on the Waymo, Small City, and Cambridge Landmarks datasets in unbounded outdoor scenes. Compared to prior methods, our approach preserves finer scene details and sharper structures, especially under large viewpoint changes, whereas previous methods often fail to preserve fine structures in rendered images.}
  \label{fig:render}
  \vspace{-0.4cm}
\end{figure*}

\subsubsection{TUGI: Uncertainty-Guided Gaussian Initialization}
TUGI encodes the geometric reliability of each 3D point into the initial state of its corresponding Gaussian primitive. For a candidate point $\mathbf{P}$ with estimated tri-view variance $\sigma_g^2$, we initialize the Gaussian center $\boldsymbol{\mu}_{\text{new}}$ at $\mathbf{P}$ and compute its color $\mathbf{c}_{\text{new}}$ as the mean of pixel intensities from all three views, ensuring appearance robustness across viewpoints. 

% To reflect geometric confidence, we scale the Gaussian covariance proportionally to the estimated standard deviation: Snew∝√σ2g\mathbf{S}_{\text{new}} \propto \sqrt{\sigma_g^2}. This allows uncertain points to begin with broader support, providing the optimizer greater flexibility in poorly constrained regions, while high-confidence points start with tighter, more localized influence.
To reflect geometric confidence, we scale the Gaussian covariance proportionally to the estimated standard deviation ($\mathbf{S}{\text{new}} \propto \sqrt{\sigma_g^2}$), which allows uncertain points to begin with broader support, providing greater flexibility to the optimizer.

We further modulate the initial opacity as a decreasing function of variance to mitigate rendering artifacts from unreliable regions. Specifically, the opacity is defined as $\alpha_{\text{new}} = \text{sigmoid}^{-1}(a(1 - k \sqrt{\sigma_g^2}))$, where $a$ is a base opacity and $k$ controls the degree of attenuation. In this way, TUGI adaptively downweights uncertain primitives during early optimization, facilitating convergence toward a compact and accurate map.

% \begin{figure*}[t]
%   \centering
%   \includegraphics[width=0.98\linewidth]{figs/render-1.png}
%   % \captionsetup{aboveskip=0pt, belowskip=0pt}
%   % \vspace{-0.2cm}
%   \caption{Rendering quality comparison on the Waymo, Small City, and Cambridge Landmarks datasets in unbounded outdoor scenes. Compared to prior methods, our approach preserves finer scene details and sharper structures, especially under large viewpoint changes, whereas previous methods often fail to preserve fine structures in rendered images.}
%   \label{fig:render}
%   % \vspace{-0.4cm}
% \end{figure*}

\section{Experiment}
\label{sec:exp}

\textbf{Datasets}:
To comprehensively evaluate our system, we conduct experiments on three challenging outdoor datasets. The \textbf{Waymo Open Dataset}~\cite{sun2020waymo} features long-range driving sequences with prolonged low-parallax motion and large texture-less regions. The \textbf{Small City}~\cite{kerbl2024hierarchical} sequences are characterized by a multitude of dynamic objects and significant illumination changes. Finally, the \textbf{Cambridge Landmarks Dataset}~\cite{ferens2023hyperpose} involves hand-held camera captures with extremely aggressive motions and frequent lighting variations. Together, these datasets systematically test the tracking robustness and mapping quality under various real-world conditions.

\textbf{Metrics}: We evaluate our system in terms of both tracking accuracy and mapping quality. For tracking performance, we use the root-mean-square error (RMSE) of the absolute trajectory error (ATE). For mapping quality, we employ three widely-used metrics: peak signal-to-noise ratio (PSNR), structural similarity index (SSIM), and learned perceptual image patch similarity (LPIPS).

\textbf{Baselines}: 
We benchmark our method against state-of-the-art neural rendering SLAM methods (RGB-only), including the NeRF-based \textbf{NICER-SLAM} \cite{zhu2024nicer} and three leading 3DGS-based methods: \textbf{Photo-SLAM} \cite{huang2023photo}, \textbf{MonoGS} \cite{matsuki2023monogs}, and \textbf{OpenGS-SLAM} \cite{wang2024opengs}.

\begin{table*}[t]
\centering	
\scriptsize
\renewcommand\arraystretch{1}
\renewcommand\tabcolsep{1.4pt}
\vspace{+0.1cm}
\caption{Evaluation on the Waymo dataset, comparing tracking accuracy (ATE RMSE [m] $\downarrow$) and mapping quality (PSNR, SSIM, LPIPS).}
\label{tab:waymo_combined}
% \resizebox{\textwidth}{!}{%
\begin{tabular}{l|rccc|rccc|rccc|cccc|cccc}
\toprule
& \multicolumn{4}{c}{\textbf{NICER-SLAM~\cite{zhu2024nicer}}}& \multicolumn{4}{c}{\textbf{Photo-SLAM~\cite{huang2023photo}}}& \multicolumn{4}{c}{\textbf{MonoGS~\cite{matsuki2023monogs}}}& \multicolumn{4}{c}{\textbf{OpenGS-SLAM~\cite{wang2024opengs}}}& \multicolumn{4}{c}{\textbf{Ours}}\\
\cmidrule(lr){2-5} \cmidrule(lr){6-9} \cmidrule(lr){10-13} \cmidrule(lr){14-17} \cmidrule(lr){18-21}
\textbf{Scene} & \textbf{ATE↓}& \textbf{PSNR↑}& \textbf{SSIM↑}& \textbf{LPIPS↓}& \textbf{ATE↓}& \textbf{PSNR↑}& \textbf{SSIM↑}& \textbf{LPIPS↓}& \textbf{ATE↓}& \textbf{PSNR↑}& \textbf{SSIM↑}& \textbf{LPIPS↓}& \textbf{ATE↓}& \textbf{PSNR↑}& \textbf{SSIM↑}& \textbf{LPIPS↓}& \textbf{ATE↓}& \textbf{PSNR↑}& \textbf{SSIM↑}& \textbf{LPIPS↓}\\
% & (m) ↓ & ↑ & ↑ &  ↓ & (m) ↓ & ↑ & ↑ & ↓ & (m) ↓ & ↑ & ↑ &  ↓ & (m) ↓ & ↑ & ↑ &  ↓ & (m) ↓ & ↑ & ↑ &  ↓ \\
\midrule
100613& 19.390& 11.46& 0.624& 0.705& 14.280& 14.29& 0.655& 0.794& 6.953& 21.89& 0.779& 0.543& 0.324& 24.41& 0.811& 0.360& \textbf{0.282}& \textbf{27.51}& \textbf{0.859}& \textbf{0.297}\\
13476& 8.180& 8.59& 0.507& 0.817& 25.850& 17.36& 0.726& 0.663& 3.366& 20.95& 0.723& 0.693& 0.422& 22.29& 0.733& 0.602& \textbf{0.406}& \textbf{22.96}& \textbf{0.742}& \textbf{0.490}\\
106762& 35.590& 10.46& 0.425& 0.670& 58.320& 18.95& 0.802& 0.558& 18.160& 22.24& 0.814& 0.515& 0.893& 26.19& \textbf{0.851}& 0.326& \textbf{0.721}& \textbf{26.19}& 0.845& \textbf{0.306}\\
132384& 25.220& 15.12& 0.782& 0.536& 3.752& 20.03& 0.839& 0.510& 12.080& 23.48& 0.856& 0.427& 0.436& 26.98& 0.883& 0.283& \textbf{0.100}& \textbf{29.45}& \textbf{0.906}& \textbf{0.247}\\
152706& 18.670& 11.55& 0.625& 0.745& 18.100& 17.92& 0.766& 0.768& 9.180& 22.52& 0.791& 0.649& 0.309& 23.95& 0.802& 0.533& \textbf{0.280}& \textbf{25.50}& \textbf{0.816}& \textbf{0.449}\\
153495& 15.420& 11.15& 0.487& 0.743& 6.407& 18.21& 0.730& 0.746& 5.718& 21.49& 0.782& 0.635& 1.576& 23.66& 0.800& 0.499& \textbf{0.870}& \textbf{25.61}& \textbf{0.834}& \textbf{0.354}\\
158686& 20.590& 12.65& 0.609& 0.756& 21.990& 16.96& 0.696& 0.684& 8.396& 21.25& \textbf{0.734}& 0.574& 1.076& 21.71& 0.731& 0.468& \textbf{0.932}& \textbf{22.07}& 0.732& \textbf{0.399}\\
163453& 22.680& 15.38& 0.690& 0.748& 25.390& 18.58& 0.739& 0.694& 11.210& 19.28& 0.743& 0.642& 1.719& 21.00& 0.745& 0.506& \textbf{1.443}& \textbf{22.76}& \textbf{0.775}& \textbf{0.397}\\
405841& 10.600& 13.66& 0.621& 0.815& 5.466& 17.31& 0.724& 0.655& 1.703& 23.14& 0.804& 0.522& 0.800& 25.72& 0.840& 0.333& \textbf{0.385}& \textbf{26.33}& \textbf{0.845}& \textbf{0.313}\\
\midrule
Average & 19.593& 12.22& 0.597& 0.726& 19.951& 17.73& 0.742& 0.675& 8.530& 21.80& 0.781& 0.578& 0.839& 23.99& 0.800& 0.434& \textbf{0.602}& \textbf{25.38}& \textbf{0.817}& \textbf{0.361}\\
\bottomrule
\end{tabular}%
% }
\vspace{-0.2cm}
\end{table*}

\textbf{Implementation Details}:
We use DUST3R \cite{wang2024dust3r} as our dense matcher and initialize each frame's pose using matches to the previous keyframe.
For keyframe selection, we follow common criteria in~\cite{matsuki2023monogs}. 
We adopt the relative scale accumulation strategy used in~\cite{wang2024opengs}, allowing us to maintain metric coherence across tri-view correspondences. 
The tracking optimization is performed for 40 iterations per frame and 80 iterations for mapping. Adam optimizer is used with learning rates of 0.001 for rotation and 0.002 for translation. The weights for the geometric losses in Eq.~\ref{eq:total_loss} are empirically set to $\lambda_{2D}=0.01$ and $\lambda_{3D}=0.01$. 
For the DART mechanism described in Eq.~\ref{eq:dart}, we set the weight bounds $w_{max}=1.0$ and $w_{min}=0.1$, with midpoint $N_m=5$ and steepness factor $k=0.8$. All experiments are conducted on a desktop computer equipped with an NVIDIA RTX 4090 GPU and an Intel Core i9-13900K CPU.

\begin{table}[t]
\centering
\scriptsize
\renewcommand\arraystretch{1}
\renewcommand\tabcolsep{6pt}
% \vspace{-0.2cm}
\caption{Evaluation on the Small City dataset.}
\label{tab:smallcity_results}
% \resizebox{\columnwidth}{!}{%
\begin{tabular}{l l c c c c}
\toprule
\textbf{Scene} & \textbf{Method} & \textbf{ATE (m) ↓}& \textbf{PSNR ↑}& \textbf{SSIM ↑}& \textbf{LPIPS ↓}\\
\midrule
\multirow{3}{*}{01} & MonoGS & 2.673& 15.92& 0.443& 0.824 \\
& OpenGS-SLAM & 6.356& 16.59& 0.456& 0.664 \\
& Ours & \textbf{0.883}& \textbf{18.70}& \textbf{0.508}& \textbf{0.516} \\
\cmidrule(l){2-6}
\multirow{3}{*}{02} & MonoGS & 3.372& 18.69& 0.583& 0.678 \\
& OpenGS-SLAM & 3.080& 18.93& 0.607& 0.472 \\
& Ours & \textbf{1.973}& \textbf{19.37}& \textbf{0.617}& \textbf{0.429} \\
\cmidrule(l){2-6}
\multirow{3}{*}{03} & MonoGS & 3.381& 17.35& 0.477& 0.684 \\
& OpenGS-SLAM & 1.006& 17.38& 0.483& 0.552 \\
& Ours & \textbf{0.728}& \textbf{18.25}& \textbf{0.538}& \textbf{0.510} \\
\midrule
\multirow{3}{*}{Average} & MonoGS & 3.142& 17.32& 0.501& 0.729 \\
& OpenGS-SLAM & 3.481& 17.63& 0.515& 0.563 \\
& Ours & \textbf{1.195}& \textbf{18.78}& \textbf{0.554}& \textbf{0.485} \\
\bottomrule
\end{tabular}%
% }
\vspace{-0.2cm}
\end{table}

\begin{table}[t]
\centering
\scriptsize
\renewcommand\arraystretch{1}
\renewcommand\tabcolsep{6pt}
\caption{Evaluation on the Cambridge Landmarks dataset.}
\label{tab:cambridge_results}
% \resizebox{\columnwidth}{!}{%
\begin{tabular}{l l c c c c}
\toprule
\textbf{Scene} & \textbf{Method} & \textbf{ATE (m) ↓}& \textbf{PSNR ↑}& \textbf{SSIM ↑}& \textbf{LPIPS ↓}\\
\midrule
\multirow{3}{*}{seq 1} & MonoGS & 11.720& 17.25& 0.705& 0.523\\
& OpenGS-SLAM & 12.926& 19.87& 0.733& 0.386 \\
& Ours & \textbf{1.762}& \textbf{20.38}& \textbf{0.748}& \textbf{0.326}\\
\cmidrule(l){2-6}
\multirow{3}{*}{seq 2} & MonoGS & 9.021& 12.32& 0.249& 0.662\\
& OpenGS-SLAM & 6.520& 13.58& 0.298& 0.502 \\
& Ours & \textbf{2.619}& \textbf{14.05}& \textbf{0.302}& \textbf{0.486}\\
\cmidrule(l){2-6}
\multirow{3}{*}{seq 3} & MonoGS & 31.894& 13.69& 0.423& 0.691\\
& OpenGS-SLAM & 2.889& 16.76& 0.534& 0.460 \\
& Ours & \textbf{1.447}& \textbf{17.63}& \textbf{0.570}& \textbf{0.404} \\
\cmidrule(l){2-6}
\multirow{3}{*}{seq 4} & MonoGS & 6.701& NAN& NAN& NAN\\
& OpenGS-SLAM & 3.624& 12.78& 0.302& 0.615 \\
& Ours & \textbf{2.209}& \textbf{13.83}& \textbf{0.332}& \textbf{0.595} \\
\midrule
\multirow{3}{*}{Average} & MonoGS & 14.834& 14.42& 0.459& 0.625\\
& OpenGS-SLAM & 6.490& 15.75& 0.467& 0.491 \\
& Ours & \textbf{2.009}& \textbf{16.47}& \textbf{0.488}& \textbf{0.453}\\
\bottomrule
\end{tabular}%
% }
% \vspace{-0.3cm}
\end{table}

\subsection{Result \& Analysis}
\begin{figure}[h]
  \centering
  \includegraphics[width=\linewidth]{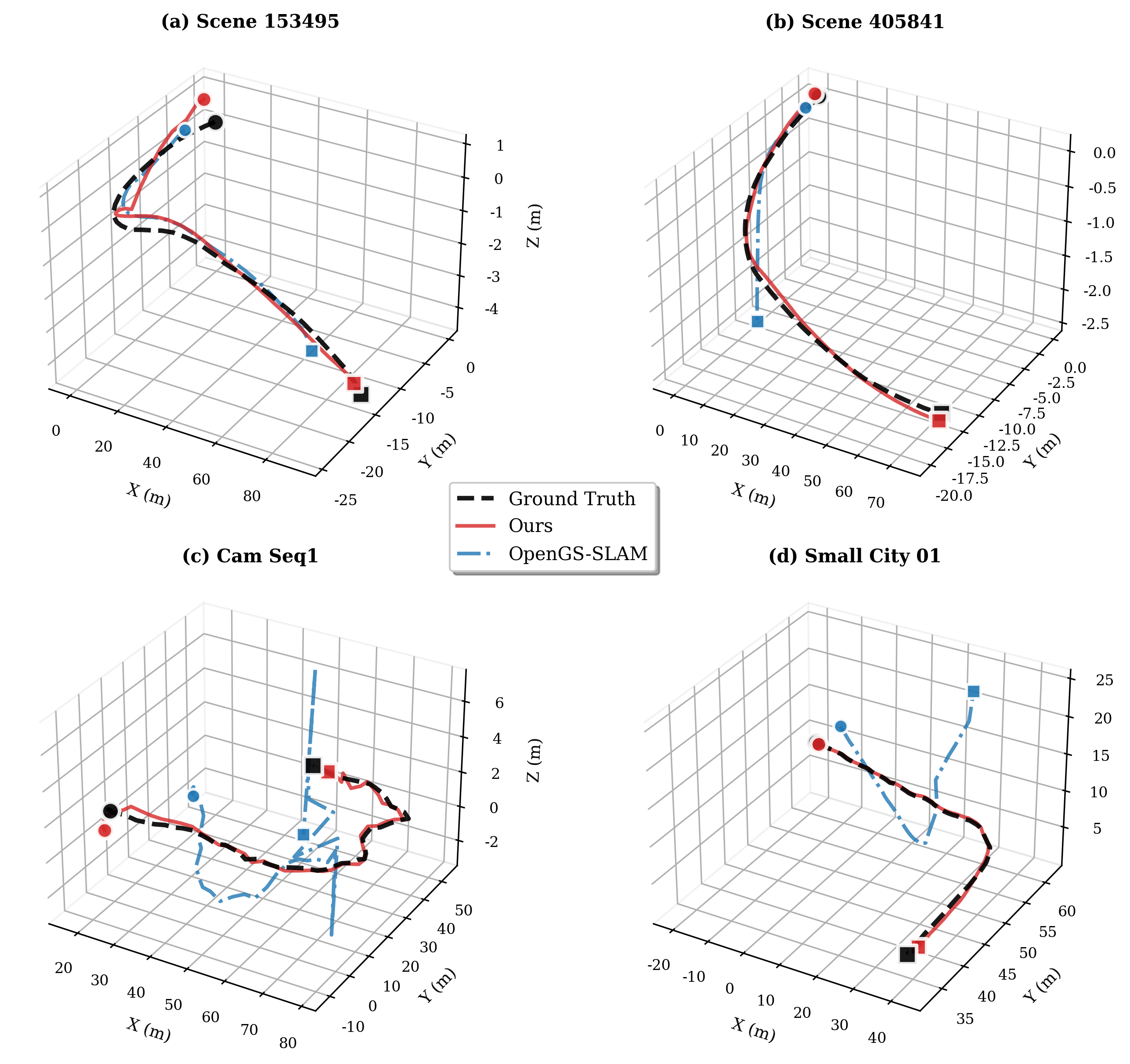}
  \caption{Trajectory comparison on challenging outdoor sequences. Black dashed lines: ground truth; red: our method; blue: OpenGS-SLAM. Our method maintains superior tracking accuracy while OpenGS-SLAM exhibits significant drift during rapid motion.}
  \label{fig:traj_comparison}
  \vspace{-0.2cm}
\end{figure}

We conducted a comprehensive evaluation of TVG-SLAM on three challenging outdoor datasets, benchmarking it against state-of-the-art GS SLAM methods. Quantitative results are detailed in \cref{tab:waymo_combined,tab:smallcity_results,tab:cambridge_results}, with qualitative comparisons illustrated in \cref{fig:render,fig:traj_comparison}. The results consistently demonstrate that our method achieves significant superiority in both tracking accuracy and mapping quality.

\textbf{Performance on the Waymo Dataset.} This dataset presents a severe test of SLAM robustness, characterized by long-range driving scenarios with low parallax and unbounded environments featuring large sky areas. As shown in \cref{tab:waymo_combined}, TVG-SLAM achieves a substantial improvement in tracking accuracy. The average ATE is reduced by approximately \textbf{28.2\%} compared to OpenGS-SLAM, and shows an order-of-magnitude advantage over MonoGS and Photo-SLAM. This accuracy boost is a direct result of our robust hybrid geometric constraints in the tracking. In low-parallax, long-straight-road scenarios, purely photometric methods are prone to significant drift, whereas our $\mathcal{L}_{2D}$ and $\mathcal{L}_{3D}$ losses provide stable geometric references that effectively suppress this drift. As illustrated by the trajectory comparison in \cref{fig:traj_comparison}, OpenGS-SLAM exhibits significant drift, whereas our trajectory remains closely aligned with the ground truth. Furthermore, this precise pose estimation provides a solid foundation for high-quality mapping, enabling us to achieve an average PSNR of \textbf{25.38}, significantly outperforming all competing methods.

\textbf{Performance on Small City \& Cambridge Landmarks.} These two datasets demand higher levels of system stability and responsiveness, characterized by numerous dynamic elements and aggressive camera motion, respectively. Our advantages become even more pronounced in these scenarios. As reported in \cref{tab:smallcity_results,tab:cambridge_results}, TVG-SLAM reduces the average ATE by \textbf{65.7\%} on Small City and \textbf{69.0\%} on Cambridge Landmarks compared to OpenGS-SLAM. This highlights the robustness of our tracking framework, where the DART mechanism plays a critical role. During aggressive motions, when mapping latency increases rendering uncertainty, DART adaptively down-weights the unreliable photometric loss, forcing the system to rely more on stable geometric constraints and preventing tracking failure. Our method also excels in mapping quality. Compared to OpenGS-SLAM, our method improves the average PSNR by \textbf{6.5\%} on Small City and \textbf{4.6\%} on Cambridge Landmarks, respectively. The qualitative results in~\cref{fig:render} visually confirm the superiority of our strategy. Compared to the blurry and artifact-ridden renderings from baseline methods, our rendering results preserve finer details and clearer geometric structures. This is attributable to our uncertainty-guided initialization, which generates more physically plausible Gaussian primitives, thereby enabling higher-fidelity mapping on top of accurate tracking.

TVG-SLAM outperforms existing methods in both low-parallax driving and dynamic hand-held scenarios, demonstrating the effectiveness of our tri-view geometry and uncertainty-guided mapping framework.

\begin{table}[t]
\centering
\scriptsize
\renewcommand\arraystretch{1.3}
\renewcommand\tabcolsep{15pt}
\caption{Ablation study of our key components on the Waymo dataset (scene: 153495). We evaluate the impact on tracking accuracy (ATE) and mapping quality (PSNR).}
\label{tab:ablation}
\begin{tabular}{lcc}
\toprule
\textbf{Components} & \textbf{ATE (m) ↓}& \textbf{PSNR ↑}\\
\midrule
\textbf{full model (Ours)} & \textbf{0.870}& \textbf{25.62}\\
\hdashline
\noalign{\vskip 0.5ex}
\multicolumn{3}{l}{\textit{Ablation on Tracking Strategy}} \\
w/o $L_{3D}$ (point-point)& 1.038& 25.23 \\
w/o $L_{2D}$ (Tri-view) & 1.193& 25.44 \\
w/o TGC ($L_{2D}$ $\&$ $L_{3D}$) & 1.269& 25.45 \\
w/o DART& 1.053& 25.31 \\
\hdashline
\noalign{\vskip 0.5ex}
\multicolumn{3}{l}{\textit{Ablation on Mapping Strategy}} \\
w/o TUGI& 1.203& 24.90\\
\bottomrule
\end{tabular}%
% \vspace{-0.3cm}
\end{table}

\subsection{Ablation Study}
In this section, we conduct a series of ablation experiments to validate the effectiveness of our proposed key components. The results on a representative Waymo scene are presented in~\cref{tab:ablation}.

\textbf{Impact of Geometric Constraints:} 
Removing the 2D tri-view trifocal loss (w/o $L_{\text{2D}}$) or the 3D alignment loss (w/o $L_{\text{3D}}$) leads to a notable increase in ATE, rising from 0.870m to 1.193m and 1.038m, respectively. 
Furthermore, removing both geometric constraints together (w/o TGC) results in the worst performance (ATE = 1.269 m), highlighting their complementary roles.
These results confirm that tri-view geometric supervision is essential for accurate and stable pose tracking, especially under challenging photometric conditions.

\begin{figure}[h]
  \centering
  \includegraphics[width=0.9\linewidth]{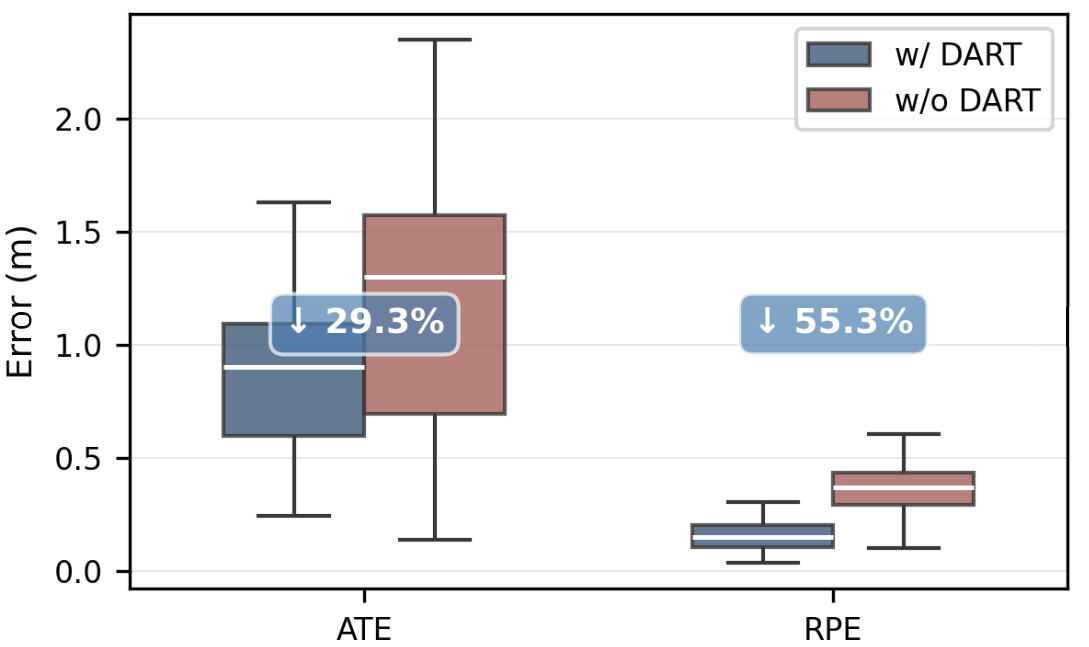}
  \caption{Ablation in DART on the Waymo dataset (scene: 158686).}
  \label{fig:dart_ablation}
  % \vspace{-0.3cm}
\end{figure}

\textbf{Impact of DART:} 
As shown in~\cref{tab:ablation}, disabling DART (w/o DART) degrades tracking performance, increasing ATE from 0.870 m to 1.053 m (↑21.0\%).
\cref{fig:dart_ablation} further shows that DART reduces both ATE and RPE by 29.3\% and 55.3\%, respectively, with lower variances.
These results validate that dynamically attenuating photometric loss during mapping staleness enables more reliance on robust geometric constraints, enhancing both the accuracy and stability of pose estimation under fast motion and scene changes.
This validates the benefit of DART in asynchronous SLAM systems, where rendering lag can otherwise compromise pose estimation.

\textbf{Impact of TUGI:} 
Among all single-component ablations, removing TUGI causes the largest degradation in tracking accuracy, with ATE increasing by 38.3\%. 
It also results in the most significant drop in rendering quality, with PSNR decreasing by 0.72dB—the steepest decline in the table.
This demonstrates that a well-initialized map is not only essential for stable pose estimation but also critical for preserving high-fidelity rendering quality.
By encoding geometric uncertainty into Gaussian shape and opacity, TUGI effectively enhances both the accuracy and physical plausibility of the 3D map.

\textbf{Efficiency Analysis.}
As shown in \Cref{tab:efficiency}, dense matching with~\cite{wang2024dust3r} accounts for most of the runtime, as it was chosen for its high geometric reliability. Our modular framework allows this to be replaced with lighter alternatives, enabling future real-time operation.

\begin{table}[t]
\centering
\vspace{+0.1cm}
\caption{Efficiency analysis of TVG-SLAM.}
\label{tab:efficiency}
\scriptsize
\renewcommand\arraystretch{0.9}
\renewcommand\tabcolsep{4.3pt}
\begin{tabular}{@{}l|ccccc|cc@{}}
\toprule
& \multicolumn{5}{c|}{\textbf{Tracking}} & \multicolumn{2}{c}{\textbf{Mapping}} \\
\cmidrule(l){2-6} \cmidrule(l){7-8}
\textbf{Stage} & matching & pose/scale & render & $L_{2D}$ & $L_{3D}$ & init/update & render \\
\midrule
\textbf{Time (ms)} & 400 & 49 & 3.5/it & 6.0/it & 7.2/it & 472 & 4.5/it \\
\bottomrule
\end{tabular}
\vspace{-0.2cm}
\end{table}


\section{CONCLUSIONS}
\label{sec:conclusion}
We presented \textbf{TVG-SLAM}, an RGB-only SLAM system that addresses the limitations of photometric-based tracking in 3DGS. By leveraging a tri-view geometry paradigm, our system introduces dense tri-view matching, Hybrid Geometric Tracking, uncertainty-guided Gaussian initialization (TUGI), and an adaptive photometric weighting mechanism (DART).
Experiments on challenging outdoor datasets demonstrate that TVG-SLAM achieves state-of-the-art performance in both tracking and mapping, especially under large viewpoint changes and dynamic lighting.
Future work will explore lightweight alternatives to dense matching, incorporate loop closure for large-scale consistency, and extend the system to dynamic environments.


%%%%%%%%%%%%%%%%%%%%%%%%%%%%%%%%%%%%%%%%%%%%%%%%%%%%%%%%%%%%%%%%%%%%%%%%%%%%%%%%

%%%%%%%%%%%%%%%%%%%%%%%%%%%%%%%%%%%%%%%%%%%%%%%%%%%%%%%%%%%%%%%%%%%%%%%%%%%%%%%%



%\section*{ACKNOWLEDGMENT}

%The preferred spelling of the word ÒacknowledgmentÓ in America is without an ÒeÓ after the ÒgÓ. Avoid the stilted expression, ÒOne of us (R. B. G.) thanks . . .Ó  Instead, try ÒR. B. G. thanksÓ. Put sponsor acknowledgments in the unnumbered footnote on the first page.

%%%%%%%%%%%%%%%%%%%%%%%%%%%%%%%%%%%%%%%%%%%%%%%%%%%%%%%%%%%%%%%%%%%%%%%%%%%%%%%%

%References are important to the reader; therefore, each citation must be complete and correct. If at all possible, references should be commonly available publications.

%\bibliographystyle{IEEEtranS}
\bibliographystyle{IEEEtran}
\bibliography{References_RAL}

\end{document}